\documentclass[utf8]{frontiersSCNS} % for Science, Engineering and Humanities and Social Sciences articles
\usepackage{url,hyperref,lineno,microtype}
\usepackage[onehalfspacing]{setspace}

\usepackage{amsthm,amsmath}
\usepackage[utf8]{inputenc} %unicode support
\usepackage{appendix}
\usepackage{amsfonts}
\usepackage{mathtools}
\usepackage{graphicx}
\DeclareGraphicsExtensions{.pdf,.png,.jpg,.mps,.eps,.ps}

\usepackage{rotating}

\usepackage{alphalph}
\renewcommand*{\thesubfigure}{\roman{subfigure}}

\usepackage{url}
\newcommand{\burl}[1]{\url{#1}} % forces to use URL for bibliography urls

\usepackage{floatrow}
\usepackage{float}

\usepackage{comment}

\usepackage[symbol]{footmisc}

\usepackage{forloop}
\newcommand{\defvec}[1]{\expandafter\newcommand\csname v#1\endcsname{{\mathbf{#1}}}}
\newcounter{ct}
\forLoop{1}{26}{ct}{
	\edef\letter{\alph{ct}}
	\expandafter\defvec\letter
}

\forLoop{1}{26}{ct}{
	\edef\letter{\Alph{ct}}
	\expandafter\defvec\letter
}

\newcommand{\norm}[1]{\ensuremath{\Vert{#1}\Vert}}

\def\keyFont{\fontsize{8}{11}\helveticabold }
\def\firstAuthorLast{Jordan {et~al.}} %use et al only if is more than 1 author
\def\Authors{Ian D. Jordan\,$^{1,3}$, Piotr Aleksander Sok\'{o}\l\,$^{2}$ and Il Memming Park\,$^{1,2,3,*}$}
\def\Address{$^{1}$Department of Applied Mathematics and Statistics, Stony Brook University, Stony Brook, NY, USA \\
$^{2}$Department of Neurobiology and Behavior, Stony Brook University, Stony Brook, NY, USA \\
$^{3}$Institute of Advanced Computational Science, Stony Brook University, Stony Brook, NY, USA}
\def\corrAuthor{Il Memming Park : Manuscript contains 7286 words and 12 figures}

\def\corrEmail{memming.park@stonybrook.edu}

\begin{document}
\onecolumn
\firstpage{1}

\title[GRUs viewed through the lens of continuous time dynamical systems]{Gated recurrent units viewed through the lens of continuous time dynamical systems} 

\author[\firstAuthorLast ]{\Authors} %This field will be automatically populated
\address{} %This field will be automatically populated
\correspondance{} %This field will be automatically populated

\extraAuth{}% If there are more than 1 corresponding author, comment this line and uncomment the next one.
%\extraAuth{corresponding Author2 \\ Laboratory X2, Institute X2, Department X2, Organization X2, Street X2, City X2 , State XX2 (only USA, Canada and Australia), Zip Code2, X2 Country X2, email2@uni2.edu}

\maketitle

\Huge
\textbf{PLEASE CITE THE PUBLISHED VERSION OF THIS MANUSCRIPT \\ DOI: 10.3389/fncom.2021.678158}
\normalsize

\begin{abstract}

\section{}
%In recent years, the efficacy of using artificial recurrent neural networks to model cortical dynamics has been a topic of interest.
Gated recurrent units (GRUs) are specialized memory elements for building recurrent neural networks.
Despite their incredible success on various tasks, including extracting dynamics underlying neural data, little is understood about the specific dynamics representable in a GRU network.
%, and how these dynamics play a part in performance and generalization.
As a result, it is both difficult to know a priori how successful a GRU network will perform on a given task, and also their capacity to mimic the underlying behavior of their biological counterparts.
Using a continuous time analysis, we gain intuition on the inner workings of GRU networks. We restrict our presentation to low dimensions, allowing for a comprehensive visualization. 
We found a surprisingly rich repertoire of dynamical features that includes stable limit cycles (nonlinear oscillations), multi-stable dynamics with various topologies, and homoclinic bifurcations.
At the same time we were unable to train GRU networks to produce continuous attractors, which are hypothesized to exist in biological neural networks.
We contextualize the usefulness of different kinds of observed dynamics and support our claims experimentally.

\tiny
 \keyFont{ \section{Keywords:} Recurrent Neural Networks, Dynamical Systems, Continuous Time, Bifurcations,Time-Series}
\end{abstract}

\section{Introduction}
Recurrent neural networks (RNNs) can capture and utilize sequential structure in natural and artificial languages, speech, video, and various other forms of time series.
The recurrent information flow within an RNN implies that the data seen in the past has influence on the current state of the RNN, forming a mechanism for having memory through (nonlinear) temporal traces that encode both \textit{what} and \textit{when}.
Past works have used RNNs to study neural population dynamics \citep{Costa2017}, and have demonstrated qualitatively similar dynamics between biological neural networks and artificial networks trained under analogous conditions \citep{mante_2013, Sussillo2015ANN, Cueva2020LowdimensionalDF}. In turn, this brings into question the efficacy of using such networks as a means to study brain function.
With this in mind, training standard vanilla RNNs to capture long-range dependences within a sequence is challenging due to the vanishing gradient problem \citep{hochreiter_untersuchungen_1991,bengio_learning_1994}.
Several special RNN architectures have been proposed to mitigate this issue, notably the long short-term memory (LSTM) units \citep{hochreiter_long_1997} which explicitly guard against unwanted corruption of the information stored in the hidden state until necessary.
Recently, a simplification of the LSTM called the \textit{gated recurrent unit} (GRU) \citep{cho_learning_2014} has become popular in the computational neuroscience and machine learning communities thanks to its performance in 
speech \citep{prabhavalkar_comparison_2017}, music \citep{choi_convolutional_2017}, video \citep{dwibedi_temporal_2018}, and extracting nonlinear dynamics underlying neural data \citep{pandarinath_inferring_2018}.
However, certain mechanistic tasks, specifically unbounded counting, come easy to LSTM networks but not to GRU networks \citep{weiss_practical_2018}.  

Despite these empirical findings, we lack systematic understanding of the internal time evolution of GRU's memory structure and its capability to represent nonlinear temporal dynamics. Such an understanding will make clear what specific tasks (natural and artificial) can or cannot be performed \citep{bengio_learning_1994}, how computation is implemented \citep{sussillo_opening_2012,Beer2006}, and help to predict qualitative behavior \citep{Zhao2016, Beer1995}. In addition, a great deal of the literature discusses the local dynamics (equilibrium points) of RNNs \citep{bengio_learning_1994, sussillo_opening_2012}, but a complete theory requires an understanding of the global properties as well \citep{Beer1995}.
Furthermore, a deterministic understanding of a GRU network's topological structure will provide fundamental insight as to a trained network's generalization ability, and therefore help in understanding how to seed RNNs for specific tasks \citep{Doya1993, 9049080}.

In general, the hidden state dynamics of an RNN can be written as
$\vh_{t+1} = f(\vh_t, \vx_t)$
where $\vx_t$ is the current input in a sequence indexed by $t$, $f$ is a nonlinear function, and $\vh_t$ represents the hidden memory state that carries all information responsible for future output.
In the absence of input, $\vh_t$ evolves over time on its own:
\begin{align}\label{eq:RNN:DS}
	\vh_{t+1} = f(\vh_t)
\end{align}
where $f(\cdot) \coloneqq f(\cdot, \mathbf{0})$ for notational simplicity.
In other words, we can consider the temporal evolution of memory stored within an RNN as a trajectory of an autonomous dynamical system defined by \eqref{eq:RNN:DS}, and use dynamical systems theory to further investigate and classify the temporal features obtainable in an RNN.
In this paper, we intend on providing a deep intuition of the inner workings of the GRU through a continuous time analysis. While RNNs are traditionally implemented in discrete time, we show in the next section that this form of the GRU can be interpreted as a numerical approximation of an underlying system of ordinary differential equations. Historically, discrete time systems are often more challenging to analyze when compared with their continuous time counterparts, primarily due to their more \textit{jumpy} nature, allowing for more complex dynamics in low-dimensions \citep{pasemann_simple_1997, DBLP:conf/iclr/0001B17}. Due to the relatively continuous nature of many abstract and physical systems, it may be of great use to analyze the underlying continuous time system of a trained RNN directly in some contexts, while interpreting the added dynamical complexity from the discretization as anomalies from numerical analysis \citep{7780459, heath_scientific_2018, leveque_numerical_1992, thomas_numerical_1995}. Furthermore, the recent development of \textit{Neural Ordinary Differential Equations} have catalyzed the computational neuroscience and machine learning communities to turn much of their attention to continuous-time implementations of neural networks \citep{NEURIPS2018_69386f6b, morrill_neural_2021}. 

We discuss a vast array of observed local and global dynamical structures, and validate the theory by training GRUs to predict time series with prescribed dynamics.  
As to not compromise the presentation, we restrict our analysis to low dimensions for easy visualization \citep{Zhao2016, Beer1995}. However, given a trained GRU of any finite dimension, the findings here still apply, and can be applied with further analysis on a case by case basis (more information on this in the discussion). Furthermore, to ensure our work is accessible we will assume a pedagogical approach in our delivery. We recommend Meiss \citep{meiss_differential_2007} for more background on the subject.

\section{Underlying Continuous Time System of Gated Recurrent Units}\label{sec:cont}
The GRU uses two internal gating variables: the \textit{update gate} $\vz_t$ which protects the $d$-dimensional hidden state $\vh_t \in \mathbb{R}^d$ and the \textit{reset gate} $\vr_t$ which allows overwriting of the hidden state and controls the interaction with the input $\vx_t \in \mathbb{R}^p$.
\begin{alignat}{5}
	\label{eq:1}
	\vz_t &= \sigma(\vW_z \vx_t + \vU_z \vh_{t-1} + \vb_z)\quad
	\\
	\label{eq:2}
	\vr_t &= \sigma(\vW_r \vx_t + \vU_r \vh_{t-1} + \vb_r)\quad
	\\
	\label{eq:3}
	\vh_t &= (1 - \vz_t) \odot \tanh(\vW_h \vx_t + \vU_h( \vr_t \odot \vh_{t-1}) + \vb_h)
	+ \vz_t \odot \vh_{t-1} \quad
\end{alignat}
where $\vW_z, \vW_r, \vW_h \in \mathbb{R}^{d \times p}$ and $\vU_z, \vU_r, \vU_h \in \mathbb{R}^{d \times d}$ are the parameter matrices, $\vb_z, \vb_r, \vb_h \in \mathbb{R}^{d}$ are bias vectors, $\odot$ represents element-wise multiplication, and $\sigma(\vz) = {1}/{(1 + e^{-\vz})}$ is the element-wise logistic sigmoid function.
Note that the hidden state is asymptotically contained within $[-1, 1]^d$ due to the saturating nonlinearities, implying that if the state is initialized outside of this trapping region, it must eventually enter it in finite time and remain in it for all later time.

Note that the update gate $\vz_t$ controls how fast each dimension of the hidden state decays, providing an adaptive time constant for memory.
Specifically, as $\lim_{\vz_t \to 1} \vh_t = \vh_{t-1}$, GRUs can implement perfect memory of the past and ignore $\vx_t$.
Hence, a $d$-dimensional GRU is capable of keeping a near constant memory through the update gate---near constant since $0 < [\vz_t]_j < 1$, where $[\cdot]_j$ denotes $j$-th component of a vector.
Moreover, the autoregressive weights (mainly $\vU_h$ and $\vU_r$) can support time evolving memory (\citep{DBLP:conf/iclr/0001B17} considered this a hindrance and proposed removing all complex dynamical behavior in a simplified GRU).

To investigate the memory structure further, let us consider the dynamics of the hidden state in the absence of input, i.e. $\vx_t = 0, \forall t$, which is of the form \eqref{eq:RNN:DS}. From a dynamical system's point of view, all inputs to the system can be understood as perturbations to the autonomous system, and therefore have no effect on the set of achievable dynamics.
To utilize the rich descriptive language of continuous time dynamical systems theory, we recognize the autonomous GRU-RNN as a weighted forward Euler discretization to the following continuous time dynamical system:
%We now recognize the GRU as a forward Euler discretization to the following continuous time dynamical system:
%To utilize the rich descriptive language of continuous time dynamical system theory, we consider the following continuous time limit of the (autonomous) GRU time evolution:
\begin{align} 
	\vz(t) &=\sigma(\vU_z \vh(t) + \vb_z)\quad
	\\
	\vr(t) &=\sigma(\vU_r \vh(t) + \vb_r)\quad
	\\
	\dot{\vh}  &= (1 - \vz(t)) \odot(
	\tanh(\vU_h(\vr(t) \odot \vh(t)) + \vb_h) -\vh(t))\quad
	\label{eq:GRU:general:continuous}
\end{align}
%where $\oslash$ denotes point-wise division and
where $\dot{\vh} \equiv \frac{\mathrm{d}\vh(t)}{\mathrm{d}t}$.
Since both $\sigma(\cdot)$ and $\tanh(\cdot)$ are smooth, this continuous limit is justified and serves as a basis for further analysis, as all GRU networks are attempting to approximate this continuous limit.
In the following, GRU will refer to the continuous time version~\eqref{eq:GRU:general:continuous}.
Note that the update gate $\vz(t)$ again plays the role of a state-dependent time constant for memory decay. We note, however, that $\vz(t)$ adjusts flow speed point-wise, resulting in non-constant nonlinear slowing of all trajectories, as $\vz(t) \in (0, 1)$.
Since $1 - \vz(t) > 0$, and thus cannot change sign, it acts as a homeomorphism between \eqref{eq:GRU:general:continuous} and the same system with this leading multiplicative term removed. Therefore, it does not change the topological structure of the dynamics \citep{kuznetsov_elements_1998}, and we can safely ignore the effects of $\vz(t)$ in the following theoretical analysis sections (\ref{sec:theory:1d} \& \ref{sec:theory:2d}). In these sections we set $\vU_z=0$ and $\vb_z=0$. A derivation of the continuous time GRU can be found in section 1 of the supplementary material. Further detail on the effects of $\vz(t)$ are discussed in the final section of this paper.

% one GRU
\section{Stability Analysis of a One Dimensional GRU}\label{sec:theory:1d}
\begin{figure}[th]
	\centering
	\includegraphics[width=\textwidth]{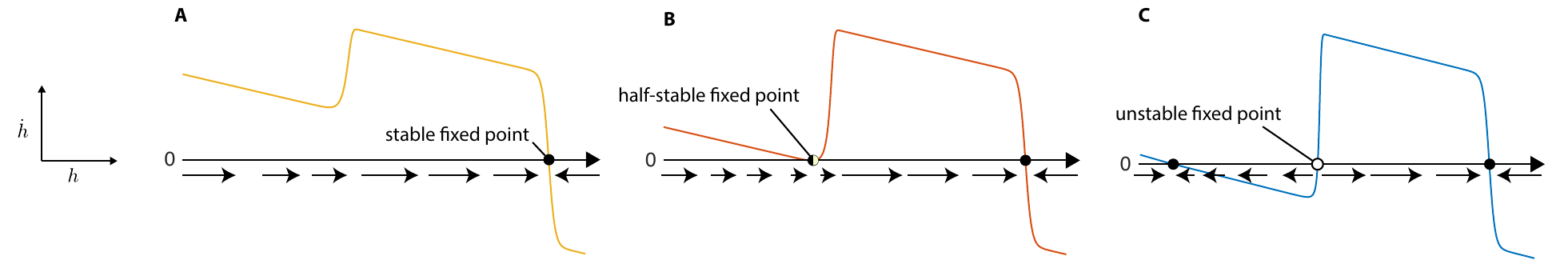}
	\caption{
		Three possible types of one dimensional flow for a 1D GRU.
		When $\dot{h} > 0$, $h(t)$ increases.
		This flow is indicated by a rightward arrow.
		Nodes ($\{h \mid \dot{h}(h) = 0\}$) are represented as circles and classified by their stability \citep{meiss_differential_2007}.
	}
	\label{fig:1D}
\end{figure}
For a 1D GRU\footnote{The number/dimension of GRUs references to the dimension of the hidden state dynamics.} ($d=1$), \eqref{eq:GRU:general:continuous} reduces to a one dimensional dynamical system where every variable is a scalar.
The expressive power of a 1D GRU is quite limited, as only three stability structures (topologies) exist (see section 2 in the supplementary material): (\textbf{A}) a single stable node, (\textbf{B}) a stable node and a half-stable node, and (\textbf{C}) two stable nodes separated by an unstable node (see Fig.~\ref{fig:1D}).
The corresponding time evolution of the hidden state are (A) decay to a fixed value, (B) decay to a fixed value, but from one direction halt at an intermediate value until perturbed, or (C) decay to one of two fixed values (bistability).
%The bistability can be used to capture a binary latent state in the input sequence.
The bistability can be used to model a switch, such as in the context of simple decision making, where inputs can perturb the system back and forth between states. 

The topology the GRU takes is determined by its parameters.
If the GRU begins in a region of the parameter space corresponding to (A), we can smoothly vary the parameters to transverse (B) in the parameter space, and end up at (C).
This is commonly known as a saddle-node bifurcation.
Speaking generally, a bifurcation is the change in topology of a dynamical system, resulting from a smooth change in parameters.
The point in parameter space at which the bifurcation occurs is called the bifurcation point (e.g. Fig.~\ref{fig:1D}B), and we will refer to the fixed point that changes its stability at the bifurcation point as the \textit{bifurcation fixed point} (e.g. the half-stable fixed point in Fig.~\ref{fig:1D}B).
The codimension of a bifurcation is the number of parameters which must vary in order to remain on the bifurcation manifold. In the case of our example, a saddle-node bifurcation is codimension-1~\citep{kuznetsov_elements_1998}.
Right before transitioning to (B), from (A), the flow near where the half-stable node would appear can exhibit arbitrarily slow flow. We will refer to these as \textit{slow points} \citep{sussillo_opening_2012}. In this context, slow points allow for metastable states, where a trajectory will flow towards the slow point, remain there for a period of time, before moving to the stable fixed point.

% Two GRUs
\section{Analysis of a Two Dimensional GRU}\label{sec:theory:2d}
\begin{figure}[t!b!hp]
	\centering
	\includegraphics[width=1\textwidth]{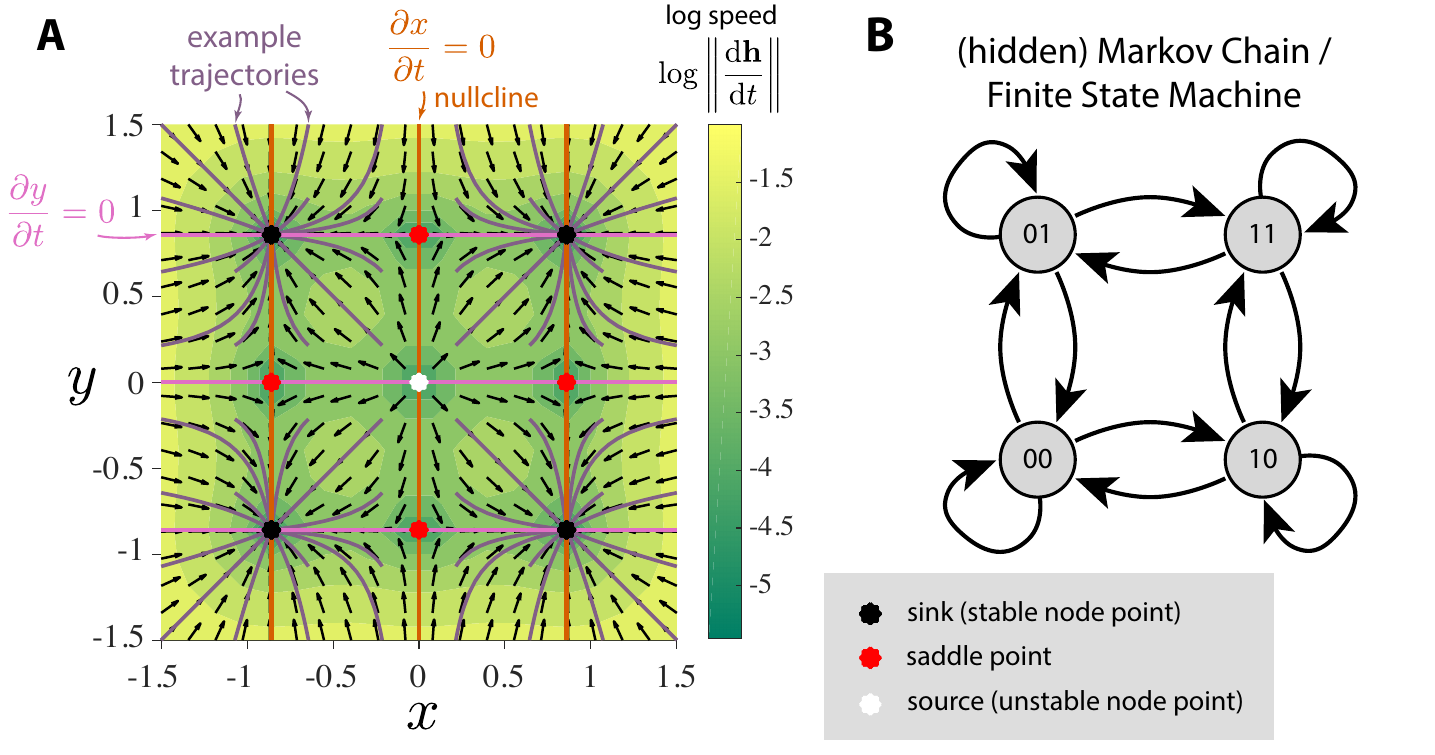}
	\caption{
		Illustrative example of two independent bistable GRUs.
		(\textbf{A}) Phase portrait.
		The flow field $\dot{\vh} = [\dot{x}, \dot{y}]^\top$ is decomposed into direction (black arrows) and speed (color).
		Purple lines represent trajectories of the hidden state which converge to one of the four stable fixed points.
		Note the four quadrants coincide with the basin of attraction for each of the stable nodes.
		The fixed points appear when the x- and y-nullclines intersect.
		(\textbf{B}) The four stable nodes of this system can be interpreted as a continuous analogue of 4-discrete states with input-driven transitions.
	}
	\label{fig:9fp:example}
\end{figure}

We will see that the addition of a second GRU opens up a substantial variety of possible topological structures.
For notational simplicity, we denote the two dimensions of $\vh$ as $x$ and $y$.
We visualize the flow fields defined by \eqref{eq:GRU:general:continuous} in 2-dimensions as \textit{phase portraits} which reveal the topological structures of interest~\citep{meiss_differential_2007}.
For starters, the phase portrait of two independent bistable GRUs can be visualized as Fig.~\ref{fig:9fp:example}A. It clearly shows 4 stable states as expected, with a total of 9 fixed points. This could be thought of as a continuous-time continuous-space implementation of a finite state machine with 4 states (Fig.~\ref{fig:9fp:example}B). The 3 types of observed fixed points---stable (sinks), unstable (sources), and saddle points---exhibit locally linear dynamics, however, the global geometry is nonlinear and their topological structures can vary depending on their arrangement.

%\begin{SCfigure}[30]%[t!b!hp]

We explored stability structures attainable by 2D GRUs. Due to the relatively large number of observed topologies, this section's main focus will be on demonstrating all observed local and global dynamical features obtainable by 2D GRUs. A catalog of all known topologies can be found in section 3 of the supplementary material, along with the parameters of every phase portrait depicted in this paper. We cannot say whether or not this catalog is exhaustive, but the sheer number of structures found is a testament to the expressive power of the GRU network, even in low dimensions.   

\begin{figure}[b!ht]
	\centering
	\includegraphics[width=\textwidth]{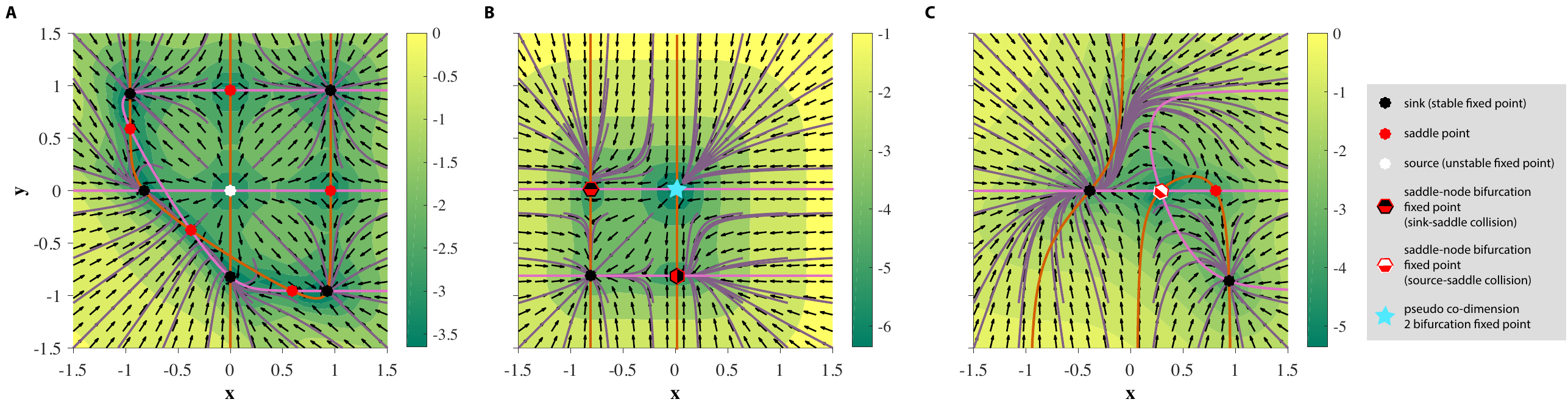}
	\caption{Existence of all observed simple fixed points and bifurcation fixed points with 2D GRUs, depicted in phase space. Orange and pink lines represent the x and y nullclines respectively. Purple lines indicate various trajectories of the hidden state. Direction of the flow is determined by the black arrows, where the colormap underlaying the figure depicts the magnitude of the velocity of the flow in log scale.}
	\label{fig:fixedpoints} \label{fig:ex} \label{fig:ex1} \label{fig:ex2} \label{fig:ex3}
\end{figure}

\begin{figure}[h]
	\floatbox[{\capbeside\thisfloatsetup{capbesideposition={right,top},capbesidewidth=8cm}}]{figure}[\FBwidth]
	{\caption{A cartoon representation of the observed \textit{pseudo-codimension-2} bifurcation fixed point. This structure occurs in implementation when placing a sink (top right), a source (bottom left), and two saddle points (top left and bottom right) close enough together, such that the distance between the two points furthest away from one another $d$ is below machine precision $\epsilon$. Under such conditions, the local dynamics behave as a hybridization of all four points. Since at least two parameters need to be adjusted in order to achieve this behavior, we give it the label of \textit{pseudo-codimension-2}; \textit{pseudo} because $d$ can never equal $0$ in this system.}\label{fig:codim2}}
	{\includegraphics[width=0.33\textwidth]{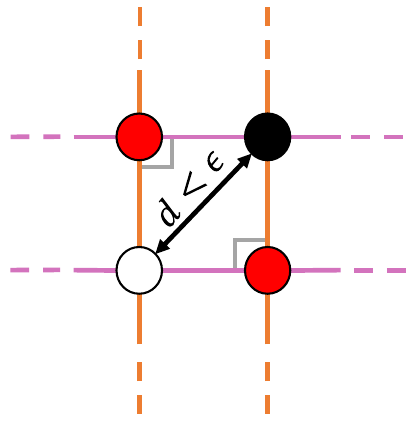}}
\end{figure}

Before proceeding, let us take this time to describe all the local dynamical features observed. In addition to the previously mentioned three types of fixed points, 2D GRUs can exhibit a variety of bifurcation fixed points, resulting from regions of parameter space that separate all topologies restricted to simple fixed points (i.e stable, unstable, and saddle points).
Behaviorally speaking, these fixed points act as hybrids between the previous three, resulting in a much richer set of obtainable dynamics.
In Fig.~\ref{fig:ex}, we show all observed types of fixed points.\footnote{
	2D GRUs feature both codimension-1 and pseudo-codimension-2 bifurcation fixed points.
	In codimension-1, we have the saddle-node bifurcation fixed point, as expected from its existence in the 1D GRU case.
	These can be thought of as both the fusion of a stable fixed point and a saddle point, and the fusion of an unstable fixed point and a saddle point.
	We will refer to these fixed points as saddle-node bifurcation fixed points of the first kind and second kind respectively.
}
While no codimension-2 bifurcation fixed points were observed in the 2D GRU system, a sort of \textit{pseudo-codimension-2} bifurcation fixed point was seen by placing a sink, source, and two saddle points sufficiently close together, such that, when implemented, all four points remain below machine precision, thereby acting as a single fixed point. Fig.~\ref{fig:codim2} further demonstrates this concept, and Fig.~\ref{fig:ex}B depicts and example. We will discuss later that this sort of pseudo-bifurcation point allows the system to exhibit \textit{homoclinic-like} behavior on a two dimensional compact set.
%All of these local structures are depicted in figure \ref{fig:ex}.
%While the existence of simple fixed points was already demonstrated (see Fig.~\ref{fig:9fp:example}A).
In Fig. \ref{fig:ex}A, we see 11 fixed points, the maximum number of fixed points observed in a 2D GRU system.
A closer look at this system reveals one interpretation as a continuous analogue of 5-discrete states with input-driven transitions, similar to that depicted in Fig. \ref{fig:9fp:example}.
This imposes a possible upper bound on the network's capacity to encode a finite set of states in this manner.
%We note that any nullcline presented in this paper that is not homotopic (can be continuously deformed to) a hyperbolic tangent function requires $\vU_r$ to be non-zero to exist in a subset of parameter space with a positive Lebesgue measure, as implied by the previous section. As such, figure \ref{fig:ex}A cannot be realized by a 2D $\tanh$-RNN, whose maximum number of fixed points are demonstrated by figure \ref{fig:9fp:example} (a generalization of the dynamics in section 3). From this we can conclude that any d-dimensional $\tanh$-RNN can have at most $3^d$ fixed points and $2^d$ stable states, as each dimension of the $\tanh$-RNN architecture is decoupled from the rest, which is strictly lower than the maximum number of states expressible by a d-dimensional GRU.   
% implying additional GRUs are needed for any Markov process modeled in this manner, requiring more than five discrete states. We conjecture that the system depicted in figure~\ref{fig:ex}A is the only eleven fixed point structure obtainable with 2D GRUs, as all observed structures containing the same number of fixed points are topologically equivalent to one another.

The addition of bifurcation fixed points opens the door to dynamically realize more sophisticated models.
Take for example the four state system depicted in Fig. \ref{fig:ex}B.
If the hidden state is set to initialize in the first quadrant of phase space (i.e $(0, \infty)^2$), the trajectory will flow towards the pseudo-codimension-2 bifurcation fixed point at the origin.
Introducing noise through the input will stochastically cause the trajectory to approach the stable fixed point at $(-1,-1)$ either directly, or by first flowing into one of the two saddle-node bifurcation fixed points of the first kind. Models of this sort can be used in a variety of applications, such as perceptual decision making \citep{wong_recurrent_2006,churchland_dynamical_2014}.
%

%% HOPF BIFURCATION! LIMIT CYCLE EXPLAINATION NEEDED BEFORE HOMOCLINIC ORBIT
We will begin our investigation into the non-local dynamics observed with 2D GRUs by showing the existence of an Andronov-Hopf bifurcation, where a stable fixed point bifurcates into an unstable fixed point surrounded by a limit cycle. A limit cycle is an attracting set with a well defined basin of attraction. However, unlike a stable fixed point, where trajectories initialized in the basin of attraction flow towards a single point, a limit cycle pulls trajectories into a stable periodic orbit. If the periodic orbit surrounds an unstable fixed point the attractor is \textit{self-exciting}, otherwise it is a \textit{hidden attractor} \citep{meiss_differential_2007}. While hidden attractors have been observed in various 2D systems, they have not been found in the 2D GRU system, and we conjecture that they do not exist. If all parameters are set to zero except for the hidden state weights, which are parameterized as a rotation matrix with an associated gain, we can introduce rotation into the vector field as a function of gain and rotation angle. Properly tuning these parameters will give rise to a limit cycle; a result of the saturating nonlinearity impeding the rotating flow velocity sufficiently distant from the origin, thereby pulling trajectories towards a closed orbit.

For $\alpha, \beta \in \mathbb{R}^+$ and $s \in \mathbb{R}$,
\begin{equation} \label{eq:8}
	\vU_z,\vU_r,\vb_z,\vb_h = 0,\; 
	\vU_h = \beta\begin{bmatrix}
		\cos{\alpha} & -\sin{\alpha} \\
		\sin{\alpha} & \cos{\alpha}
	\end{bmatrix}, \vb_r = s 
\end{equation}

\begin{figure}[t!hp]
	\centering
	\includegraphics[width=\textwidth]{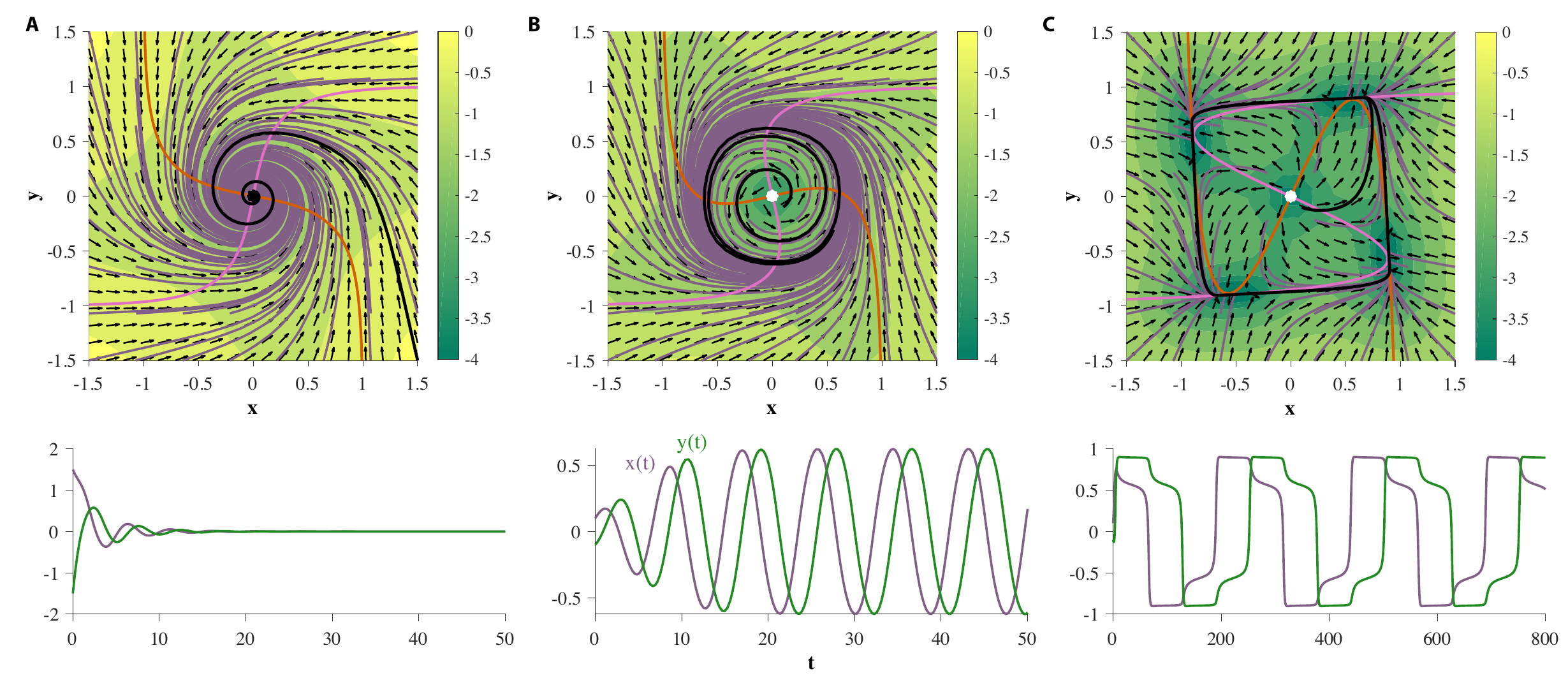}
	\caption{Two GRUs exhibit an Andronov-Hopf bifurcation, where the parameters are defined by \eqref{eq:8}. When $\alpha = \frac{\pi}{3}$ the system exhibits a single stable fixed point at the origin (Fig. \ref{fig:bifurcation}A). If $\alpha$ decreases continuously, a limit cycle emerges around the fixed point, and the fixed point changes stability (Fig. \ref{fig:bifurcation}B). Allowing $\alpha$ to decrease further increases the size and orbital period of the limit cycle (Fig. \ref{fig:bifurcation}C). The bottom row represents the hidden state as a function of time, for a single trajectory (denoted by black trajectories in each corresponding phase portrait).}
	\label{fig:bifurcation}
\end{figure}

\begin{figure}[t!hp]
	\centering
	\includegraphics[width=\textwidth]{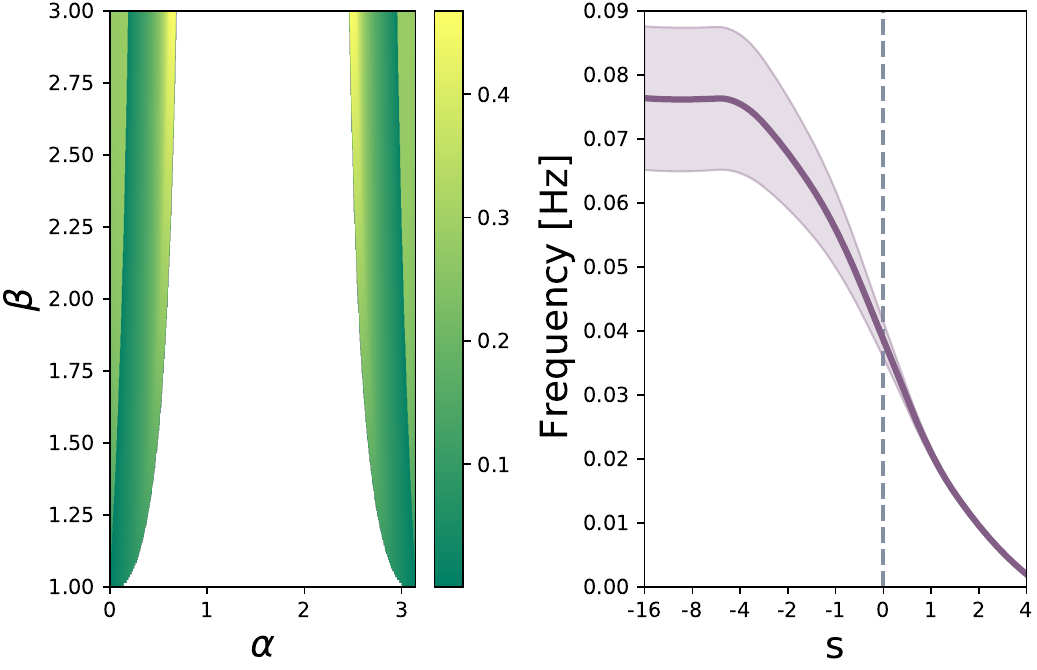}
	\caption{(Fig. \ref{fig:lastfig}-Left) parameter sweep of \eqref{eq:8} over $\alpha \in (0, \pi)$ (rotation matrix angle) and $\beta \in (1, 3)$ (gain term), for $s=0$. Color map indicates oscillation frequency in Hertz, where white space shows parameter combinations where no limit cycle exists. (Fig. \ref{fig:lastfig}-Right) average oscillation frequency across regions of the displayed $\alpha-\beta$ parameter space where a limit cycle exists. The purple shaded region depicts variance of oscillation frequency. Increasing $s$ slows down the average frequency of the limit cycles, while simultaneously reducing variance.}
	\label{fig:lastfig}
\end{figure}

Let $\beta = 3$ and $s=0$. If $\alpha = \frac{\pi}{3}$, the system has a single stable fixed point (stable spiral), as depicted in Fig. \ref{fig:bifurcation}A. If we continuously decrease $\alpha$, the system undergoes an Andronov-Hopf bifurcation at approximately $\alpha = \frac{\pi}{3.8}$. As $\alpha$ continuously  decreases, the orbital period increases, and as the nullclines can be made arbitrarily close together, the length of this orbital period can be set arbitrarily. Fig. \ref{fig:bifurcation}B shows an example of a relatively short orbital period, and Fig. \ref{fig:bifurcation}C depicts the behavior seen for slower orbits. If we continue allowing $\alpha$ to decrease, the system will undergo four simultaneous saddle-node bifurcations, and end up in a state topologically equivalent to that depicted in Fig. \ref{fig:9fp:example}A.
Fig. \ref{fig:lastfig}-Left depicts regions of the parameter space of \eqref{eq:GRU:general:continuous} parameterized by \eqref{eq:8}, where the Andronov-Hopf bifurcation manifolds can be clearly seen. Fig.\ref{fig:lastfig}-Right demonstrates one effect the reset gate can have on the frequency of the oscillations. If we alter the bias vector $b_r$, the expected oscillation period changes for regions of the $\alpha-\beta$ parameter space which exhibit a limit cycle. 
Computationally speaking, limit cycles are a common dynamical structure for modeling neuron bursting \citep{izhikevich_dynamical_2007}, taking place in many foundational works including the Hodgkin-Huxley model \citep{hodgkin_quantitative_1952} and the FitzHugh-Nagumo Model \citep{fitzhugh_impulses_1961}. Such dynamics also arise in various population level dynamics in artificial tasks, such as sine wave generation \citep{sussillo_opening_2012}. Furthermore, initializing the hidden state matrix $U_h$ of an even dimensional continuous-time RNN (tanh or GRU) with $2\times2$ blocks along the diagonal and zeros everywhere else is theoretically shown to aid in learning long-term dependencies, when all the blocks act as decoupled oscillators \citep{9049080}.

%% HOMOCLINIC BIFURCATION ADDITION

\begin{figure}[h]
	\floatbox[{\capbeside\thisfloatsetup{capbesideposition={right,top},capbesidewidth=4cm}}]{figure}[\FBwidth]
	{\caption{A 2D GRU parameterized by \eqref{eq:hcb} expresses a homoclinic orbit when $\gamma = 0.054085$ (denoted by a black trajectory). Trajectories initialized on the homoclinic orbit will approach the same fixed point in both forward and backward time.}\label{fig:homoclinic1}}
	{\includegraphics[width=0.4\textwidth]{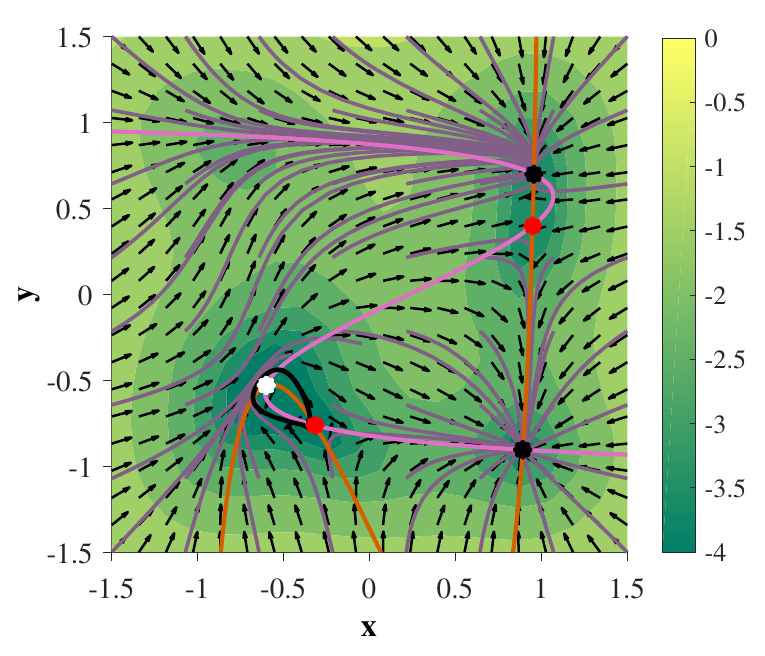}}
\end{figure}

\begin{figure}[h!]
	\centering
	\includegraphics[width=0.85\textwidth]{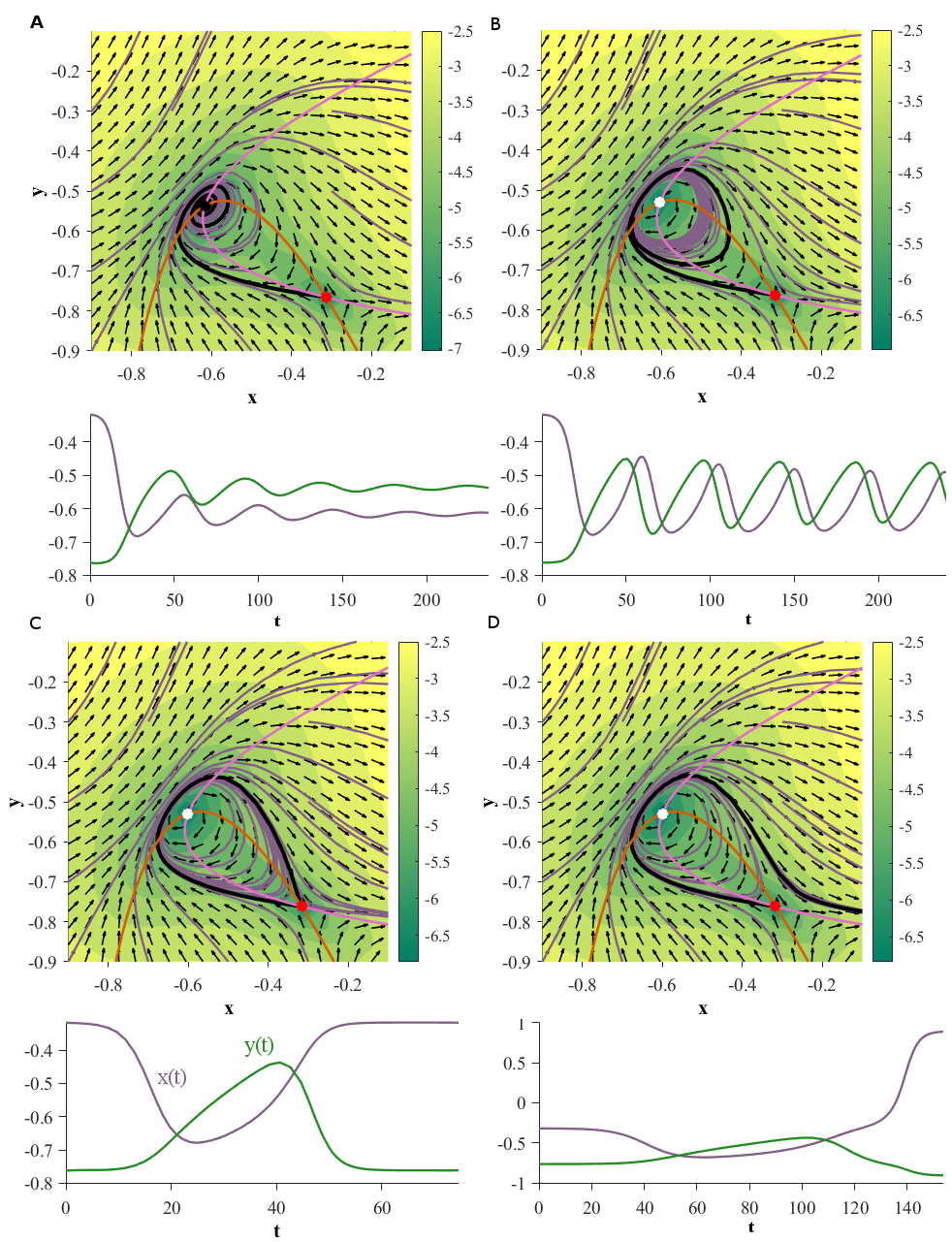}
	\caption{Two GRUs exhibit an Andronov-Hopf bifurcation followed by a homoclinic bifurcation under the same parameterization. The plots directly under each phase portrait depict the time evolution of the black trajectory for the corresponding system.  \ref{fig:homoclinicBifurcation}A ($\gamma=0.051$): the system exhibits a stable fixed point. \ref{fig:homoclinicBifurcation}B ($\gamma=0.0535$): the system has undergone an Andronov-Hopf bifurcation and exhibits a stable limit cycle. \ref{fig:homoclinicBifurcation}C ($\gamma=0.054085$): the limit cycle collides with the saddle point, creating a homoclinic orbit. \ref{fig:homoclinicBifurcation}D ($\gamma=0.0542$): the system has undergone a homoclinic bifurcation exhibits neither a homoclinic orbit nor a limit cycle.}
	\label{fig:homoclinicBifurcation}
\end{figure}

Regarding the second non-local dynamical feature, it can be shown that a 2D GRU can undergo a homoclinic bifurcation, where a periodic orbit (in this case a limit cycle) expands and collides with a saddle at the bifurcation point. At this bifurcation point the system exhibits a homoclinic orbit, where trajectories initialized on the orbit fall into the same fixed point in both forward and backward time. In order to demonstration this behavior, let the parameters of the network be defined as follows:

For $\gamma \in \mathbb{R}$,
\begin{equation} \label{eq:hcb}
	\vU_z,\vU_r,\vb_z,\vb_r = 0,\; 
	\vU_h = 3\begin{bmatrix}
		\cos{\frac{\pi}{20}} & \sin{\frac{\pi}{20}} \\
		-\sin{\frac{\pi}{20}} & \cos{\frac{\pi}{20}}
	\end{bmatrix}
	, \vb_h = \begin{bmatrix}
		0.32 \\
		\gamma
	\end{bmatrix}
\end{equation}
Under this parameterization the 2D GRU system exhibits a homoclinic orbit when $\gamma = 0.054085$ (Fig \ref{fig:homoclinic1}). 
%See section 3 of the supplementary material for a more detailed walk through of the observed homoclinic bifurcation as we adjust $\gamma$.
In order to showcase this bifurcation as well as the previous Andronov-Hopf bifurcation sequentially in action we turn to Fig \ref{fig:homoclinicBifurcation}, where the parameters are defined by \eqref{eq:hcb} and $\gamma$ is initialized at $0.051$ in Fig \ref{fig:homoclinicBifurcation}A.
% Under this selection of parameters the system exhibits a stable-fixed point (Fig \ref{fig:homoclinicBifurcation}A), whereby increasing $\gamma$ causes the system to undergo an Andronov-Hopf bifurcation, giving rise to a limit cycle (Fig \ref{fig:homoclinicBifurcation}B). As $\gamma$ continues to increase the limit cycle will grow and eventually collide with the adjacent saddle creating a homoclinic orbit out of the limit cycle (Fig \ref{fig:homoclinicBifurcation}C). Allowing $\gamma$ to increase any further will result in the annihilation of the homoclinic orbit (Fig \ref{fig:homoclinicBifurcation}D).

\begin{figure}[t!h!b]
	\centering
	\includegraphics[width=\textwidth]{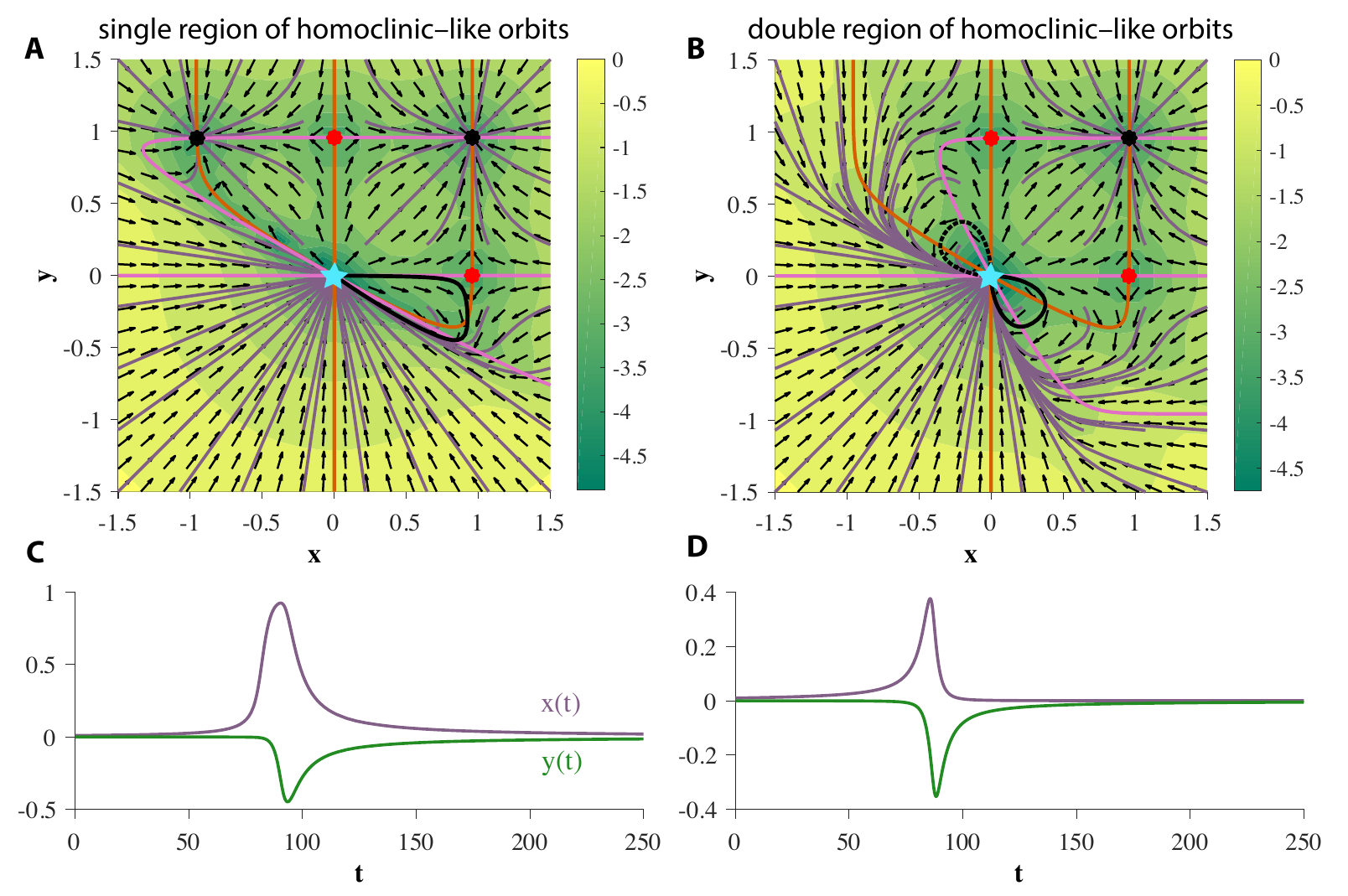}
	\caption{Two GRUs exhibit 2D bounded regions of homoclinic-like behavior. \ref{fig:homoclinic}C and \ref{fig:homoclinic}D represent the hidden state as a function of time for a single initial condition within the homoclinic-like region(s) of the single and double homoclinic-like region cases respectively (denoted by solid black trajectories in each corresponding phase portrait).}
	\label{fig:homoclinic}
\end{figure}

In addition to proper homoclinic orbits, we observe that 2D GRUs can exhibit one or two bounded planar regions of homoclinic-like orbits for a given set of parameters, as shown in Fig. \ref{fig:homoclinic}A and \ref{fig:homoclinic}B respectively. Any trajectory initialized in one of these regions will flow into the pseudo-codimension-2 bifurcation fixed point at the origin, regardless of which direction time flows in. Since the the pseudo-codimension-2 bifurcation fixed point is technically a cluster of four fixed points, including one source and one sink, as demonstrated in Fig. \ref{fig:codim2}, there is actually no homoclinic loop. However, due to the close proximity of these fixed points, trajectories repelled away from the source, but within the basin of attraction of the sink, will appear homoclinic due to the use of finite precision. This featured behavior enables the accurate depiction of various models, including neuron spiking \citep{izhikevich_dynamical_2007}.    
%

\begin{figure}[!h!]
	\centering
	\includegraphics[width=\textwidth]{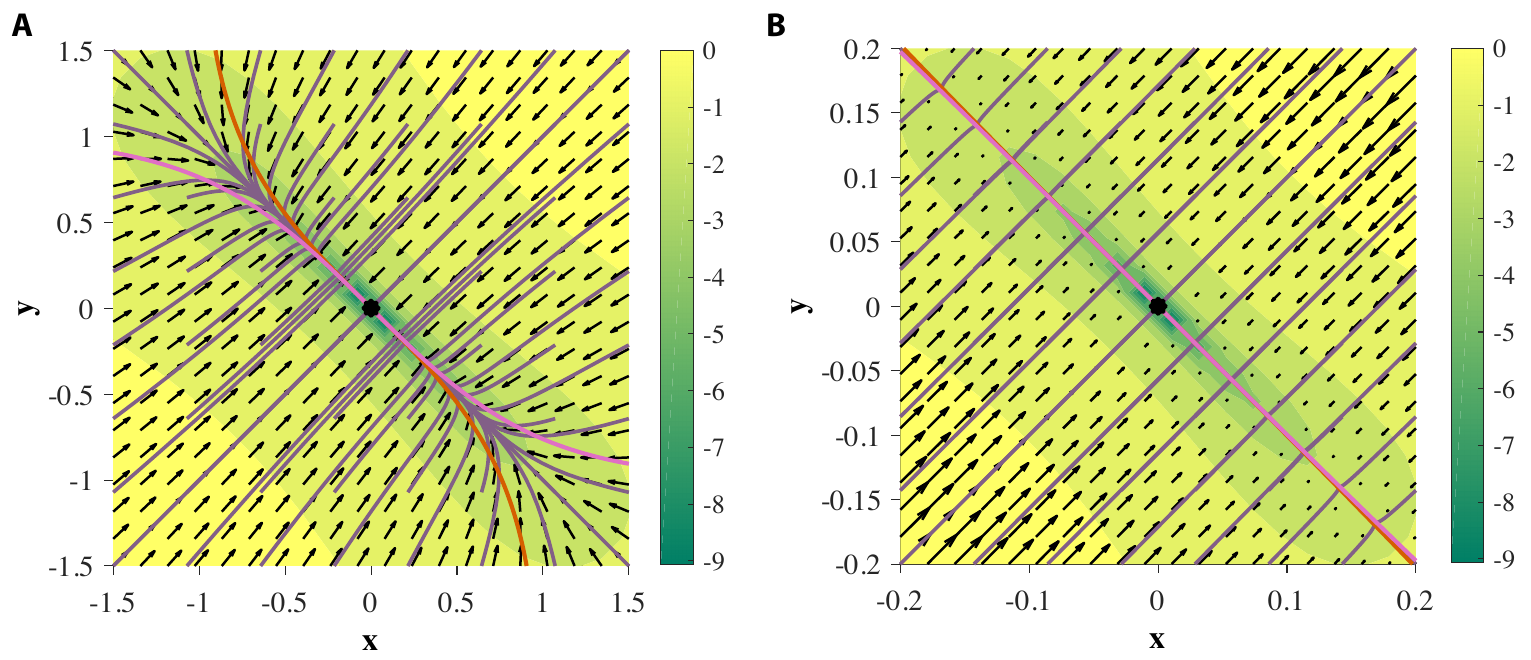}
	\caption{Two GRUs exhibit a pseudo-line attractor. Nullclines intersect at one point, but are close enough on a finite region to mimic an analytic line attractor in practice. \ref{fig:lineAttractor}A and \ref{fig:lineAttractor}B depict the same phase portrait on $[-1.5,1.5]^2$ and $[-0.2,0.2]^2$ respectively.}
	\label{fig:lineAttractor}
\end{figure}

%% LINE ATTRACTOR
With finite-fixed point topologies and global structures out of the way, the next logical question to ask is \textit{can 2D GRUs exhibit an infinite number of fixed points?} Such behavior is often desirable in models that require stationary attraction to non-point structures, such as line attractors and ring attractors. Computationally, movement along a line attractor may be interpreted as integration \citep{mante_2013}, and has been shown as a crucial population level mechanism in various tasks, including sentiment analysis \citep{maheswaranathan_reverse_2019} and decision making \citep{mante_2013}. In a similar light, movement around a ring attractor my computationally represent either modular integration or arithmetic. One known application of ring attractor dynamics in neuroscience is a representation of heading direction \citep{kim_ring_2017}.
While such behavior in the continuous GRU system has yet to be seen, an approximation of a line attractor can be made, as depicted in Fig. \ref{fig:lineAttractor}. We will refer to this phenomenon as a \textit{pseudo-line attractor}, where the nullclines remain sufficiently close on a small finite interval, thereby allowing for arbitrarily slow flow, by means of slow points. 

%Experiments
\section{Experiments: Time-Series Prediction}
As a means to put our theory to practice, in this section we explore several examples of time series prediction of continuous time planar dynamical systems using 2D GRUs.
Results from the previous section indicate what dynamical features can be learned by this RNN, and suggest cases by which training will fail.
All of the following computer experiments consist of an RNN, by which the hidden layer is made up of a 2D GRU, followed by a linear output layer.
The network is trained to make a 29-step prediction from a given initial observation, and no further input through prediction. %, 
As such, to produce accurate predictions, the RNN must rely solely on the hidden layer dynamics.

We train the network to minimize the following multi-step loss function:
\begin{equation}\label{eq:MSE}
	\mathcal{L}(\theta)
	%\mathbf{\hat{w}},\mathbf{w}
	= \frac{1}{T} \sum_{i=1}^{N_\text{traj}} \sum^{T}_{k=1}{\norm{
			\mathbf{\hat{w}}_i(k;\mathbf{w}_i(0)) - \mathbf{w}_i(k)}^2_2
	}
\end{equation}
where $\theta$ are the parameters of the GRU and linear readout, $T=29$ is the prediction horizon, $\vw_i(t)$ is the $i$-th time series generated by the true system, and $\hat{\vw}(k; \vw_0)$ is the $k$-step prediction given $\vw_0$.

\begin{figure}[t!h!b]
	\centering
	\includegraphics[width=\textwidth]{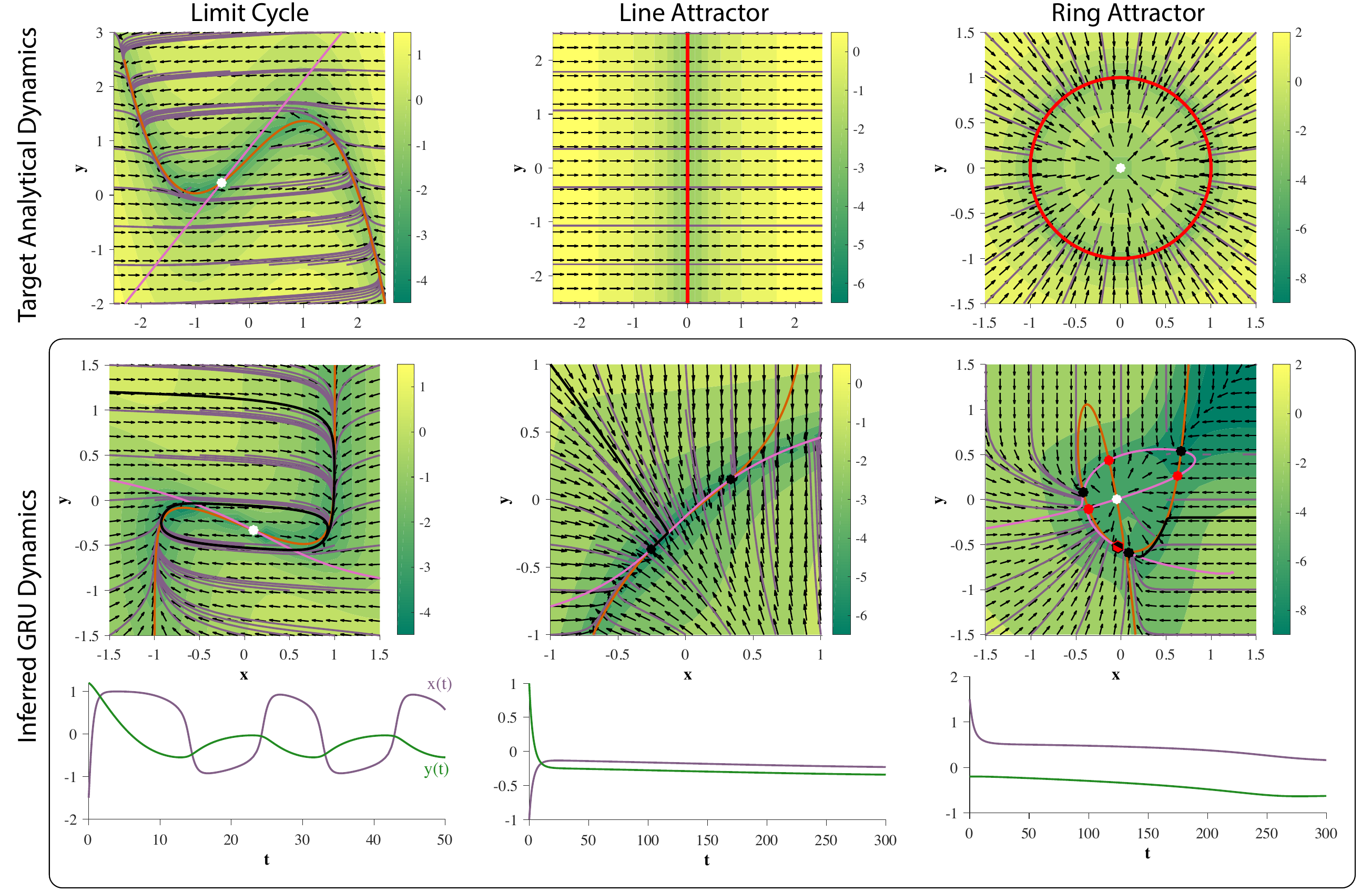}
	\caption{
		Training 2D GRUs.
		(top row) Phase portraits of target dynamical systems. Red solid lines represent 1-dimensional attractors. See main text for each system.
		(middle row) GRU dynamics learned from corresponding 29-step forecasting tasks. 
		The prediction is an affine transformation of the hidden state.
		(bottom row) An example time series generated through closed-loop prediction of the trained GRU (denoted by a black trajectory).
		GRU fails to learn the ring attractor.
	}
	\label{fig:experiments}
\end{figure}

The hidden states are initialized at zero for each trajectory.
The RNN is then trained for 4000 epochs, using ADAM \citep{kingma_adam:_2014} in whole batch mode to minimize the loss function, i.e., the mean square error between the predicted trajectory and the data.
$N_\text{traj} = 667$ time series were used for training.
Fig.~\ref{fig:experiments} depicts the experimental results of the RNN's attempt at learning each dynamical system we describe below.

\begin{figure}[t!b!hp]
	\centering
	\includegraphics[width=\textwidth]{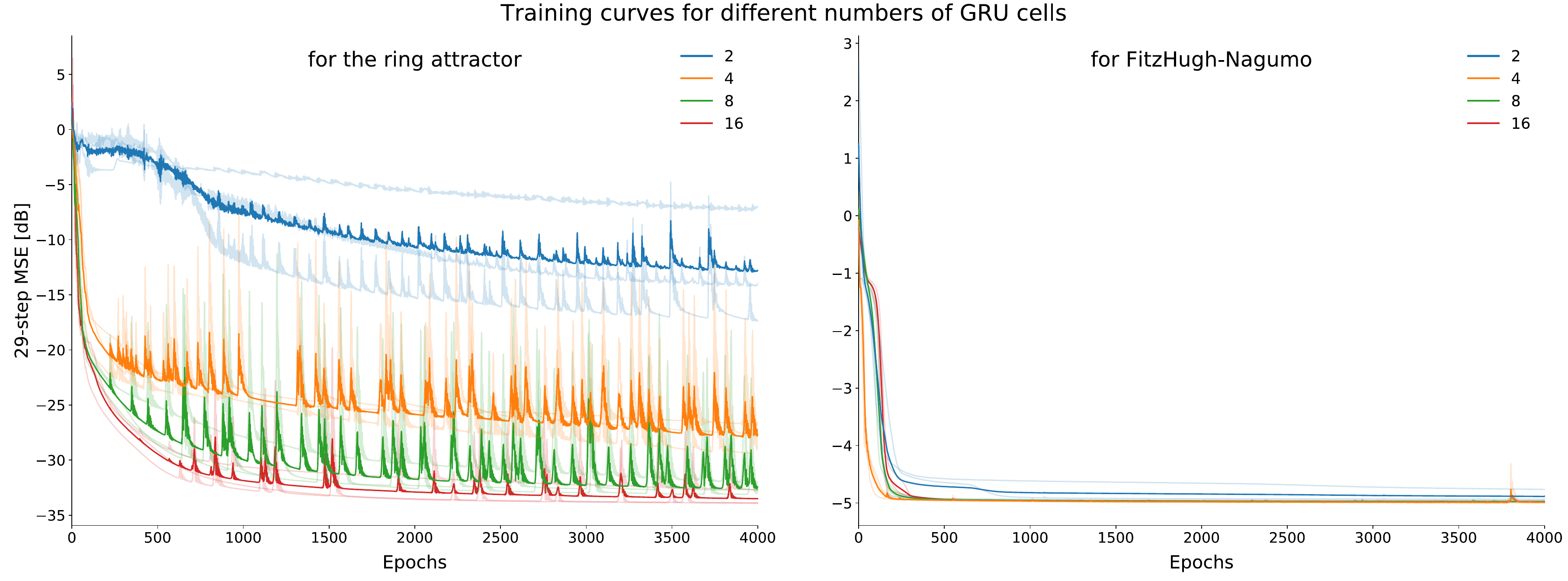}
	\caption{
		Average learning curves (training loss) for ring attractor (left) and the FitzHugh-Nagumo (right) dynamics.
		Note that the performance of the ring attractor improves as the dimensionality of the GRU increases unlike the FHN dynamics.
		Four network sizes (2, 4, 8, 16 dimensional GRU) were trained 3 times with different initializations, depicted by the more lightly colored curves.
		%Initial learning rate for ADAM was tuned using Bayesian optimization procedure.
	}
	\label{fig:learning:curves}
\end{figure}

\subsection{Limit Cycle}
To test if 2D GRUs can learn a limit cycle, we use a simple nonlinear oscillator called the FitzHugh-Nagumo Model \citep{fitzhugh_impulses_1961}.
The FitzHugh-Nagumo model is defined by:
%
%\begin{equation}
$\dot{x} = x - \frac{x^{3}}{3} - y + I_{\text{ext}}, \; \tau\dot{y} = x + a - by $,
%\end{equation}
%
where in this experiment we will chose $\tau = 12.5$, $a = 0.7$, $b = 0.8$, and $I_\text{ext} = \mathcal{N}(0.7,0.04)$.
Under this choice of model parameters, the system will exhibit an unstable fixed point (unstable spiral) surrounded by a limit cycle (Fig.~\ref{fig:experiments}).
As shown in section 4, 2D GRUs are capable of representing this topology.
The results of this experiment verify this claim (Fig.~\ref{fig:experiments}), as 2D GRUs can capture topologically equivalent dynamics.

\subsection{Line Attractor}
As discussed in section 4, 2D GRUs can exhibit a pseudo-line attractor, by which the system mimics an analytic line attractor on a small finite domain.
%sufficiently under computational constraints.
We will use the simplest representation of a planar line attractor:
%
%\begin{equation} \label{eq:line1}
$\dot{x} = -x, \quad \dot{y} = 0$.
%\end{equation}
%
This system will exhibit a line attractor along the $y$-axis, at $x=0$ (Fig.~\ref{fig:experiments}).
Trajectories will flow directly perpendicular towards the attractor.
white Gaussian noise $\mathcal{N}(0,0.1 I)$ in the training data.
%Figure \ref{fig:experiments} shows the result of the inferred dynamics after training.
While the hidden state dynamics of the trained network do not perfectly match that of an analytic line attractor, there exists a small subinterval near each of the fixed points acting as a pseudo-line attractor (Fig.~\ref{fig:experiments}).
As such, the added affine transformation (linear readout) can scale and reorient this subinterval on a finite domain. Since all attractors in a d-dimensional GRU are bound to $[-1, 1]^d$, no line attractor can extend infinitely in any given direction, which matches well with the GRUs inability to perform unbounded counting, as the continuous analog of such a task would require a trajectory to move along such an attractor. 

\subsection{Ring Attractor}
For this experiment, a dynamical system representing a standard ring attractor of radius one is used:
%
%\begin{equation} \label{eq:ring}
$
\dot{x} = -(x^{2} + y^{2} - 1)x;\; % \quad
\dot{y} = -(x^{2} + y^{2} - 1)y
$.
%\end{equation}
%
%\eqref{eq:ring1} and \eqref{eq:ring2} define the analytic model.
This system exhibits an attracting ring, centered around an unstable fixed point.
We added Gaussian noise $\mathcal{N}(0,0.1I)$ in the training data.

In our analysis we did not observe two GRUs exhibit this set of dynamics, and the results of this experiment, demonstrated in Fig. \ref{fig:experiments}, suggest they cannot.
Rather, the hidden state dynamics fall into an observed finite fixed point topology (see case xxix in section 3 of the supplementary material).
In addition, we robustly see this over multiple initializations, and the quality of approximation improves as the dimensionality of GRU increases (Fig.~\ref{fig:learning:curves}), suggesting that many GRUs are required to obtain a sufficient approximation of this set of dynamics for a practical task \citep{funahashi_approximation_1993}.

\section{Discussion}

Through example and experiment we indicated classes of dynamics which are crucial in expressing various known neural computations and obtainable with the 2D GRU network. We demonstrated the system's inability to learn continuous attractors, seemingly in any finite dimension, a structure hypothesized to exist in various neural representations. While the GRU network was not originally made as a neuroscientific model, there has been considerable work done showing high qualitative similarity between the underlying dynamics of neural recordings and artificial RNNs on the population level \citep{mante_2013, Sussillo2015ANN}.
Furthermore, recent research has modified such artificial models to simulate various neurobiological phenomenon \citep{heeger_oscillatory_2019}. One recent study demonstrated that trained RNNs of different architectures and nonlinearities express very similar fixed point topologies to one another when successfully trained on the same tasks \citep{NEURIPS2019_5f5d4720}, suggesting a possible connection in the dynamics of artificial networks and neural population dynamics. As such, an understanding of the obtainable dynamical features in a GRU network allow one to comment on the efficacy of using such an architecture as an analog of brain dynamics at the population level.
 
Although this manuscript simplified the problem by considering the 2D GRU, a lot of research has resulted in interpreting cortical dynamics as low dimensional continuous time dynamical systems \citep{zhao_variational_2020, harvey_choice-specific_2012, macdowell_low-dimensional_2020, flesch_rich_2021, mante_2013, Cueva2020LowdimensionalDF}. This is not to say that most standard neuroscience inspired tasks can be solved with such a low dimensional network. However, demonstrating that common dynamical features in neuroscience can arise in low dimensions can aid in one's ability to comment on attributes of large networks. These attributes include features such as sparsity of synaptic connections. For example, spiking models exhibiting sparse connectivity have been shown to perform comparatively with fully connected RNNs \citep{bellec_long_2018}. Additionally, pruning (i.e. removing) substantial percentages of synaptic connections in a trained RNN is known to often result in little to no drop in the network's performance on the task it was trained on \citep{frankle_lottery_2018}. This suggests two more examinable properties of large networks. The first is redundancy or multiple realizations of the dynamical mechanisms needed to enact a computation existing within the same network. For example, if only one limit cycle is sufficient to accurately perform a desired task, a trained network may exhibit multiple limit cycles, each qualitatively acting identically towards the overall computation. The second is the robustness of each topological structure to synaptic perturbation/pruning. For example, if we have some dynamical structure, say a limit cycle, how much can we move around in parameter space while still maintaining the existence of that structure?

In a related light, the GRU architecture has been used within more complex machine learning setups to interpret the real-time dynamics of neural recordings \citep{willett_high-performance_2021, pandarinath_inferring_2018}. These tools allow researchers to better understand and study the differences between neural responses, trial to trial.
Knowledge of the inner workings and expressive power of GRU networks can only further our understanding of the limitations and optimization of such setups by the same line of reasoning previously stated, thereby helping to advance this class of technologies, aiding the field of neuroscience as a whole.  

The most compared RNN architecture to the GRU is LSTM, as GRU was designed as both a model and computational simplification of this preexisting design in discrete time implementation. LSTM, for a significant period of time, was arguably the most popular discrete time RNN architecture, outperforming other models of the time on many benchmark tasks. However, there is one caveat when comparing the continuous time implementations of LSTM and GRU. A one dimensional LSTM (i.e. a single LSTM unit) is a two dimensional dynamical system, as information is stored in both the system's hidden state and cell state \citep{hochreiter_long_1997}. With the choice of analysis we use to dissect the GRU in this paper, LSTM is a vastly different class of system. We would expect to see a different and more limited array of dynamics for an LSTM unit when compared with the 2D GRU. However, we wouldn't consider this a fair comparison.  

One attribute of the GRU architecture we chose to disregard in this manuscript was the influence of the update gate $\vz(t)$. As stated in section \ref{sec:cont}, every element of this gate is bound to $(0, 1)^d$. Since \eqref{eq:GRU:general:continuous} only has one term containing the update gate, $(1 - \vz(t))$, which can be factored out, the fixed point topology does not depend on $\vz(t)$, as this term is always strictly positive. The role this gate plays is to adjust the point-wise speed of flow, and therefore can bring rise to slow manifolds. Because each element of $\vz(t)$ can become arbitrarily close to the value of one, regions of phase space associated with an element of the update-gate sufficiently close to one will experience seemingly no motion in the directions associated with those elements. For example, in the 2D GRU system, if the first element of $\vz(t)$ is sufficiently close to one, the trajectory will maintain a near fixed value in $x$. 
These slow points are not actual fixed points. Therefore, in the autonomous system, trajectories traversing them will eventually overcome this stoppage given sufficient time. However, this may add one complicating factor for analyzing implemented continuous time GRUs in practice. The use of finite precision allows for the flow speed to dip below machine precision, essentially creating \textit{pseudo-attractors} in these regions. The areas of phase space containing these points will qualitatively behave as attracting sets, but not by traditional dynamical systems terms, making them more difficult to analyze. If needed, we recommend looking at $\vz(t)$ separately, because this term acts independently from the remaining terms in the continuous time system. Therefore, any slow points found can be superimposed with the traditional fixed points in phase space.
In order to avoid the effects of finite precision all together, the system can be realized through a hardware implementation \citep{e22050537}. However, proper care needs to be given in order to mitigate analog imperfections.

Unlike the update gate, we demonstrated that the reset gate $\vr(t)$ affects the network's fixed point topology, allowing for more complicated classes of dynamics, including homoclinic-like orbits. These effects are best described through the shape of the nullclines. We will keep things qualitative here as to help build intuition. In 2D, if every element of the reset gate weight matrix $\vU_r$ and bias $\vb_r$ is zero, nullclines can  form two shapes. First is a \textit{sigmoid-like} shape (Fig. \ref{fig:bifurcation}A, \ref{fig:lineAttractor}, and \ref{fig:experiments} (inferred limit cycle and line attractor)), allowing them to intersect a line (or hyperplane in higher dimensions) orthogonal to their associated dimension a single time. The second is an \textit{s-like} shape (Fig. \ref{fig:bifurcation}B, \ref{fig:bifurcation}C, \ref{fig:homoclinic1}, and \ref{fig:experiments} (limit cycle)), allowing them to intersect a line orthogonal to their associated dimension up to three times. The peak and trough of the s-like shape can be stretched infinitely as well (Fig. \ref{fig:9fp:example}A). In this case, two fo the three resultant seemingly disconnected nullclines associated with a given dimension can be placed arbitrarily close together (Fig. \ref{fig:ex}B). Varying $\vr(t)$ allows the geometry of the nullclines to take on several additional shapes. The first of these additional structures is a \textit{pitchfork-like} shape (Fig. \ref{fig:ex}A, \ref{fig:ex}C, \ref{fig:homoclinic}). By disconnecting two of the \textit{prongs} from the pitchfork we get our second structure, simultaneously exhibiting a sigmoid-like shape and a \textit{U-like} shape (Fig. \ref{fig:ex}C). Bending the ends of the "U" at infinity down into $\mathbb{R}^2$ connects them, forming our third structure, an \textit{O-like} shape (Fig. \ref{fig:experiments} (inferred ring attractor -- orange nullcline)). This O-like shape can then also intersect the additional segment of the nullcline, creating one continuous curve (Fig. \ref{fig:experiments} (inferred ring attractor -- pink nullcline). One consequence of the reset-gate is the additional capacity to encode information in the form of stable fixed points. If we neglect $\vr(t)$, we can obtain up to four sinks (Fig. \ref{fig:9fp:example}A), as we are limited to the intersections of the nullclines; two sets of three parallel lines. Incorporating $\vr(t)$ increases the number of fixed points obtainable (Fig. \ref{fig:ex}A). Refer to section 3 of the supplementary material to see how these nullcline structures lead to a vast array of different fixed point topologies. 

Several interesting extensions to this work immediately come to mind. For one, the extension to a 3D continuous time GRU network opens up the door for the possibility of more complex dynamical features. Three spatial dimensions are the minimum required to experience chaotic dynamics in nonlinear systems \citep{meiss_differential_2007}, and due to the vast size of the GRU parameter space, even in low dimensions, such behavior is probable. Similarly, additional types of bifurcations may be present, including bifurcations of limit cycles, allowing for more complex oscillatory behavior \citep{kuznetsov_elements_1998}. Furthermore, higher dimensional GRUs may bring rise to complex center manifolds, requiring center manifold reduction to better analyze and interpret the phase space dynamics \citep{carr_applications_1981}. While we considered the underlying GRU topology separate from training, considering how the attractor structure influences learning can bring insight into successfully implementing RNN models \citep{9049080}. As of yet, this topic of research is mostly uncharted. We believe such findings, along with the work presented in this manuscript, will unlock new avenues of research on the trainability of recurrent neural networks and help to further understand their mathematical parallels with biological neural networks.

%\section{Conclusion}
%Gated neural networks were designed and used primarily as memory devices, but the temporal computation aspect is also important.
%Recurrent neural networks with gated structure have been used in a wide array of tasks, both natural and artificial. These include, but are not limited to, the interpretation of cortical circuit dynamics \citep{Costa2017} and also the design of canonical circuits using spiking neural networks~\citep{Heeger2018,Bellec2018}. 
%Understanding not only the memory structure, but what neural networks can compute over time is critical for extending the horizon of recurrent neural networks.
%Our analysis provides an intuition for how GRUs express dynamic behavior in continuous time, and reveals the rich but limited classes of dynamics the GRU can approximate in one and two dimensions. We show that 2D GRUs can exhibit a variety of expressive dynamical features, such as limit cycles, uncountably many homoclinic-like orbits, and a substantial catalog of stability structures and bifurcations. These claims were then experimentally verified. 
%We believe these findings also unlock new avenues of research on the trainability of recurrent neural networks and help to further understand their mathematical parallels with biological neural networks. 

\section*{Conflict of Interest Statement}
%All financial, commercial or other relationships that might be perceived by the academic community as representing a potential conflict of interest must be disclosed. If no such relationship exists, authors will be asked to confirm the following statement: 

The authors declare that the research was conducted in the absence of any commercial or financial relationships that could be construed as a potential conflict of interest.

\section*{Author Contributions}

I.D.J. performed the analysis, I.D.J, P.A.S, and I.M.P wrote the manuscript. P.A.S. performed the numerical experiments. I.M.P. conceived the idea, advised, and edited the manuscript. All authors read and approved the final manuscript.

\section*{Funding}
This work was supported by NIH EB-026946, and NSF IIS-1845836. I.D.J. was supported partially by the Institute of Advanced Computational Science Jr. Researcher Fellowship, at Stony Brook University.

\section*{Acknowledgments}
The authors thank Josue Nassar, Brian O'Donnell, David Sussillo, Aminur Rahman, Denis Blackmore, Braden Brinkman, Yuan Zhao, and D.S for helpful feedback and conversations regarding the analysis and writing of this manuscript.

%\section*{Data Availability Statement}
%The datasets [GENERATED/ANALYZED] for this study can be found in the [NAME OF REPOSITORY] [LINK].
% Please see the availability of data guidelines for more information, at https://www.frontiersin.org/about/author-guidelines#AvailabilityofData

%\bibliographystyle{frontiersinSCNS_ENG_HUMS} % for Science, Engineering and Humanities and Social Sciences articles, for Humanities and Social Sciences articles please include page numbers in the in-text citations
\bibliographystyle{frontiersinHLTH&FPHY} % for Health, Physics and Mathematics articles
\bibliography{grubib}

\begin{thebibliography}{49}
\expandafter\ifx\csname natexlab\endcsname\relax\def\natexlab#1{#1}\fi
\expandafter\ifx\csname urlstyle\endcsname\relax
  \expandafter\ifx\csname doi\endcsname\relax
  \def\doi#1{doi:\discretionary{}{}{}#1}\fi \else
  \expandafter\ifx\csname doi\endcsname\relax
  \def\doi{doi:\discretionary{}{}{}\begingroup \urlstyle{rm}\Url}\fi \fi
\expandafter\ifx\csname selectlanguage\endcsname\relax
  \def\selectlanguage#1{}\fi

\bibitem[{Costa et~al.(2017)Costa, Assael, Shillingford, de~Freitas, and
  Vogels}]{Costa2017}
Costa R, Assael IA, Shillingford B, de~Freitas N, Vogels T.
\newblock Cortical microcircuits as gated-recurrent neural networks.
\newblock Guyon I, Luxburg UV, Bengio S, Wallach H, Fergus R, Vishwanathan S,
  et~al., editors, {\em Advances in Neural Information Processing Systems 30\/}
  (Curran Associates, Inc.) (2017), 272--283.

\bibitem[{Mante et~al.(2013)Mante, Sussillo, Shenoy, and Newsome}]{mante_2013}
Mante V, Sussillo D, Shenoy K, Newsome W.
\newblock Context-dependent computation by recurrent dynamics in prefrontal
  cortex.
\newblock {\em Nature\/} {\bf 503} (2013) 78--84.

\bibitem[{Sussillo et~al.(2015)Sussillo, Churchland, Kaufman, and
  Shenoy}]{Sussillo2015ANN}
Sussillo D, Churchland M, Kaufman MT, Shenoy K.
\newblock A neural network that finds a naturalistic solution for the
  production of muscle activity.
\newblock {\em Nature Neuroscience\/} {\bf 18} (2015) 1025--1033.

\bibitem[{Cueva et~al.(2020)Cueva, Saez, Marcos, Genovesio, Jazayeri, Romo
  et~al.}]{Cueva2020LowdimensionalDF}
Cueva CJ, Saez A, Marcos E, Genovesio A, Jazayeri M, Romo R, et~al.
\newblock Low-dimensional dynamics for working memory and time encoding.
\newblock {\em Proceedings of the National Academy of Sciences\/} {\bf 117}
  (2020) 23021 -- 23032.

\bibitem[{Hochreiter(1991)}]{hochreiter_untersuchungen_1991}
Hochreiter S.
\newblock {\em Untersuchungen zu dynamischen neuronalen {Netzen}\/}.
\newblock Ph.D. thesis, TU Munich (1991).
\newblock Advisor J. Schmidhuber.

\bibitem[{Bengio et~al.(1994)Bengio, Simard, and
  Frasconi}]{bengio_learning_1994}
Bengio Y, Simard P, Frasconi P.
\newblock Learning long-term dependencies with gradient descent is difficult.
\newblock {\em IEEE Transactions on Neural Networks\/} {\bf 5} (1994) 157--166.
\newblock \doi{10.1109/72.279181}.

\bibitem[{Hochreiter and Schmidhuber(1997)}]{hochreiter_long_1997}
Hochreiter S, Schmidhuber J.
\newblock Long {Short}-{Term} {Memory}.
\newblock {\em Neural Computation\/} {\bf 9} (1997) 1735--1780.
\newblock \doi{10.1162/neco.1997.9.8.1735}.

\bibitem[{Cho et~al.(2014)Cho, van Merrienboer, Gulcehre, Bahdanau, Bougares,
  Schwenk et~al.}]{cho_learning_2014}
Cho K, van Merrienboer B, Gulcehre C, Bahdanau D, Bougares F, Schwenk H, et~al.
\newblock Learning {Phrase} {Representations} using {RNN} {Encoder}-{Decoder}
  for {Statistical} {Machine} {Translation}.
\newblock {\em arXiv:1406.1078 [cs, stat]\/}  (2014).
\newblock ArXiv: 1406.1078.

\bibitem[{Prabhavalkar et~al.(2017)Prabhavalkar, Rao, Sainath, Li, Johnson, and
  Jaitly}]{prabhavalkar_comparison_2017}
Prabhavalkar R, Rao K, Sainath TN, Li B, Johnson L, Jaitly N.
\newblock A {Comparison} of {Sequence}-to-{Sequence} {Models} for {Speech}
  {Recognition}.
\newblock {\em Interspeech 2017\/} (ISCA) (2017), 939--943.
\newblock \doi{10.21437/Interspeech.2017-233}.

\bibitem[{Choi et~al.(2017)Choi, Fazekas, Sandler, and
  Cho}]{choi_convolutional_2017}
Choi K, Fazekas G, Sandler M, Cho K.
\newblock Convolutional recurrent neural networks for music classification.
\newblock {\em 2017 {IEEE} {International} {Conference} on {Acoustics},
  {Speech} and {Signal} {Processing} ({ICASSP})\/} (2017), 2392--2396.
\newblock \doi{10.1109/ICASSP.2017.7952585}.

\bibitem[{Dwibedi et~al.(2018)Dwibedi, Sermanet, and
  Tompson}]{dwibedi_temporal_2018}
Dwibedi D, Sermanet P, Tompson J.
\newblock Temporal {Reasoning} in {Videos} {Using} {Convolutional} {Gated}
  {Recurrent} {Units} (2018), 1111--1116.

\bibitem[{Pandarinath et~al.(2018)Pandarinath, O'Shea, Collins, Jozefowicz,
  Stavisky, Kao et~al.}]{pandarinath_inferring_2018}
Pandarinath C, O'Shea DJ, Collins J, Jozefowicz R, Stavisky SD, Kao JC, et~al.
\newblock Inferring single-trial neural population dynamics using sequential
  auto-encoders.
\newblock {\em Nature Methods\/} {\bf 15} (2018) 805--815.
\newblock \doi{10.1038/s41592-018-0109-9}.

\bibitem[{Weiss et~al.(2018)Weiss, Goldberg, and Yahav}]{weiss_practical_2018}
Weiss G, Goldberg Y, Yahav E.
\newblock On the {Practical} {Computational} {Power} of {Finite} {Precision}
  {RNNs} for {Language} {Recognition}.
\newblock {\em arXiv:1805.04908 [cs, stat]\/}  (2018).
\newblock ArXiv: 1805.04908.

\bibitem[{Sussillo and Barak(2012)}]{sussillo_opening_2012}
Sussillo D, Barak O.
\newblock Opening the black box: Low-dimensional dynamics in high-dimensional
  recurrent neural networks.
\newblock {\em Neural Computation\/} {\bf 25} (2012) 626--649.
\newblock \doi{10.1162/NECO_a_00409}.

\bibitem[{Beer(2006)}]{Beer2006}
Beer RD.
\newblock Parameter space structure of continuous-time recurrent neural
  networks.
\newblock {\em Neural Comput.\/} {\bf 18} (2006) 3009--3051.

\bibitem[{Zhao and Park(2016)}]{Zhao2016}
Zhao Y, Park IM.
\newblock Interpretable nonlinear dynamic modeling of neural trajectories.
\newblock {\em Advances in Neural Information Processing Systems ({NIPS})\/}
  (2016).

\bibitem[{Beer(1995)}]{Beer1995}
Beer RD.
\newblock On the dynamics of small {Continuous-Time} recurrent neural networks.
\newblock {\em Adapt. Behav.\/} {\bf 3} (1995) 469--509.

\bibitem[{Doya(1993)}]{Doya1993}
Doya K.
\newblock Bifurcations of recurrent neural networks in gradient descent
  learning.
\newblock {\em IEEE Trans. Neural Netw.\/}  (1993).

\bibitem[{{Sokół} et~al.(2019){Sokół}, {Jordan}, {Kadile}, and
  {Park}}]{9049080}
{Sokół} PA, {Jordan} I, {Kadile} E, {Park} IM.
\newblock Adjoint dynamics of stable limit cycle neural networks.
\newblock {\em 2019 53rd Asilomar Conference on Signals, Systems, and
  Computers\/} (2019), 884--887.
\newblock \doi{10.1109/IEEECONF44664.2019.9049080}.

\bibitem[{Pasemann(1997)}]{pasemann_simple_1997}
Pasemann F.
\newblock A simple chaotic neuron.
\newblock {\em Physica D: Nonlinear Phenomena\/} {\bf 104} (1997) 205--211.
\newblock \doi{10.1016/S0167-2789(96)00239-4}.

\bibitem[{Laurent and von Brecht(2017)}]{DBLP:conf/iclr/0001B17}
Laurent T, von Brecht J.
\newblock A recurrent neural network without chaos.
\newblock {\em 5th International Conference on Learning Representations, {ICLR}
  2017, Toulon, France, April 24-26, 2017, Conference Track Proceedings\/}
  (OpenReview.net) (2017).

\bibitem[{He et~al.(2016)He, Zhang, Ren, and Sun}]{7780459}
He K, Zhang X, Ren S, Sun J.
\newblock Deep residual learning for image recognition.
\newblock {\em 2016 IEEE Conference on Computer Vision and Pattern Recognition
  (CVPR)\/} (2016), 770--778.
\newblock \doi{10.1109/CVPR.2016.90}.

\bibitem[{Heath(2018)}]{heath_scientific_2018}
Heath MT.
\newblock {\em Scientific {Computing}: {An} {Introductory} {Survey}, {Revised}
  {Second} {Edition}\/} (Philadelphia: SIAM - Society for Industrial and
  Applied Mathematics), second edition edn. (2018).

\bibitem[{LeVeque and Leveque(1992)}]{leveque_numerical_1992}
LeVeque RJ, Leveque R.
\newblock {\em Numerical {Methods} for {Conservation} {Laws}\/} (Basel ;
  Boston: Birkhäuser), 2nd edition edn. (1992).

\bibitem[{Thomas(1995)}]{thomas_numerical_1995}
Thomas JW.
\newblock {\em Numerical {Partial} {Differential} {Equations}: {Finite}
  {Difference} {Methods}\/} (New York: Springer), 1st ed. 1995. corr. 2nd
  printing 1998 edition edn. (1995).

\bibitem[{Chen et~al.(2018)Chen, Rubanova, Bettencourt, and
  Duvenaud}]{NEURIPS2018_69386f6b}
Chen RTQ, Rubanova Y, Bettencourt J, Duvenaud DK.
\newblock Neural ordinary differential equations.
\newblock Bengio S, Wallach H, Larochelle H, Grauman K, Cesa-Bianchi N, Garnett
  R, editors, {\em Advances in Neural Information Processing Systems\/} (Curran
  Associates, Inc.) (2018), vol.~31.

\bibitem[{Morrill et~al.(2021)Morrill, Salvi, Kidger, Foster, and
  Lyons}]{morrill_neural_2021}
Morrill J, Salvi C, Kidger P, Foster J, Lyons T.
\newblock Neural {Rough} {Differential} {Equations} for {Long} {Time} {Series}.
\newblock {\em arXiv:2009.08295 [cs, math, stat]\/}  (2021).
\newblock ArXiv: 2009.08295.

\bibitem[{Meiss(2007)}]{meiss_differential_2007}
Meiss J.
\newblock {\em Differential {Dynamical} {Systems}\/}.
\newblock Mathematical {Modeling} and {Computation} (Society for Industrial and
  Applied Mathematics) (2007).
\newblock \doi{10.1137/1.9780898718232}.

\bibitem[{Kuznetsov(1998)}]{kuznetsov_elements_1998}
Kuznetsov YA.
\newblock {\em Elements of Applied Bifurcation Theory (2nd {Ed}.)\/} (Berlin,
  Heidelberg: Springer-Verlag) (1998).

\bibitem[{Wong and Wang(2006)}]{wong_recurrent_2006}
Wong KF, Wang XJ.
\newblock A recurrent network mechanism of time integration in perceptual
  decisions.
\newblock {\em The Journal of Neuroscience: The Official Journal of the Society
  for Neuroscience\/} {\bf 26} (2006) 1314--1328.
\newblock \doi{10.1523/JNEUROSCI.3733-05.2006}.

\bibitem[{Churchland and Cunningham(2014)}]{churchland_dynamical_2014}
Churchland MM, Cunningham JP.
\newblock A {Dynamical} {Basis} {Set} for {Generating} {Reaches}.
\newblock {\em Cold Spring Harbor Symposia on Quantitative Biology\/} {\bf 79}
  (2014) 67--80.
\newblock \doi{10.1101/sqb.2014.79.024703}.

\bibitem[{Izhikevich(2007)}]{izhikevich_dynamical_2007}
Izhikevich EM.
\newblock {\em Dynamical systems in neuroscience\/} (MIT press) (2007).

\bibitem[{Hodgkin and Huxley(1952)}]{hodgkin_quantitative_1952}
Hodgkin AL, Huxley AF.
\newblock A quantitative description of membrane current and its application to
  conduction and excitation in nerve.
\newblock {\em The Journal of Physiology\/} {\bf 117} (1952) 500--544.
\newblock \doi{10.1113/jphysiol.1952.sp004764}.

\bibitem[{FitzHugh(1961)}]{fitzhugh_impulses_1961}
FitzHugh R.
\newblock Impulses and {Physiological} {States} in {Theoretical} {Models} of
  {Nerve} {Membrane}.
\newblock {\em Biophysical Journal\/} {\bf 1} (1961) 445--466.
\newblock \doi{10.1016/S0006-3495(61)86902-6}.

\bibitem[{Maheswaranathan et~al.(2019{\natexlab{a}})Maheswaranathan, Williams,
  Golub, Ganguli, and Sussillo}]{maheswaranathan_reverse_2019}
Maheswaranathan N, Williams A, Golub MD, Ganguli S, Sussillo D.
\newblock Reverse engineering recurrent networks for sentiment classification
  reveals line attractor dynamics.
\newblock {\em arXiv:1906.10720 [cs, stat]\/}  (2019{\natexlab{a}}).
\newblock ArXiv: 1906.10720.

\bibitem[{Kim et~al.(2017)Kim, Rouault, Druckmann, and
  Jayaraman}]{kim_ring_2017}
Kim SS, Rouault H, Druckmann S, Jayaraman V.
\newblock Ring attractor dynamics in the {Drosophila} central brain.
\newblock {\em Science\/} {\bf 356} (2017) 849--853.
\newblock \doi{10.1126/science.aal4835}.
\newblock Publisher: American Association for the Advancement of Science
  Section: Reports.

\bibitem[{Kingma and Ba(2014)}]{kingma_adam:_2014}
Kingma DP, Ba J.
\newblock Adam: {A} {Method} for {Stochastic} {Optimization}.
\newblock {\em arXiv:1412.6980 [cs]\/}  (2014).
\newblock ArXiv: 1412.6980.

\bibitem[{Funahashi and Nakamura(1993)}]{funahashi_approximation_1993}
Funahashi Ki, Nakamura Y.
\newblock Approximation of dynamical systems by continuous time recurrent
  neural networks.
\newblock {\em Neural Networks\/} {\bf 6} (1993) 801--806.
\newblock \doi{10.1016/S0893-6080(05)80125-X}.

\bibitem[{Heeger and Mackey(2019)}]{heeger_oscillatory_2019}
Heeger DJ, Mackey WE.
\newblock Oscillatory recurrent gated neural integrator circuits ({ORGaNICs}),
  a unifying theoretical framework for neural dynamics.
\newblock {\em Proceedings of the National Academy of Sciences\/} {\bf 116}
  (2019) 22783--22794.
\newblock \doi{10.1073/pnas.1911633116}.
\newblock Publisher: National Academy of Sciences Section: Biological Sciences.

\bibitem[{Maheswaranathan et~al.(2019{\natexlab{b}})Maheswaranathan, Williams,
  Golub, Ganguli, and Sussillo}]{NEURIPS2019_5f5d4720}
Maheswaranathan N, Williams A, Golub M, Ganguli S, Sussillo D.
\newblock Universality and individuality in neural dynamics across large
  populations of recurrent networks.
\newblock {\em Advances in Neural Information Processing Systems\/} (Curran
  Associates, Inc.) (2019{\natexlab{b}}), vol.~32.

\bibitem[{Zhao and Park(2020)}]{zhao_variational_2020}
Zhao Y, Park IM.
\newblock Variational {Online} {Learning} of {Neural} {Dynamics}.
\newblock {\em Frontiers in Computational Neuroscience\/} {\bf 14} (2020).
\newblock \doi{10.3389/fncom.2020.00071}.
\newblock Publisher: Frontiers.

\bibitem[{Harvey et~al.(2012)Harvey, Coen, and
  Tank}]{harvey_choice-specific_2012}
Harvey CD, Coen P, Tank DW.
\newblock Choice-specific sequences in parietal cortex during a
  virtual-navigation decision task.
\newblock {\em Nature\/} {\bf 484} (2012) 62--68.
\newblock \doi{10.1038/nature10918}.
\newblock Number: 7392 Publisher: Nature Publishing Group.

\bibitem[{MacDowell and Buschman(2020)}]{macdowell_low-dimensional_2020}
MacDowell CJ, Buschman TJ.
\newblock Low-{Dimensional} {Spatiotemporal} {Dynamics} {Underlie}
  {Cortex}-wide {Neural} {Activity}.
\newblock {\em Current Biology\/} {\bf 30} (2020) 2665--2680.e8.
\newblock \doi{10.1016/j.cub.2020.04.090}.

\bibitem[{Flesch et~al.(2021)Flesch, Juechems, Dumbalska, Saxe, and
  Summerfield}]{flesch_rich_2021}
Flesch T, Juechems K, Dumbalska T, Saxe A, Summerfield C.
\newblock Rich and lazy learning of task representations in brains and neural
  networks.
\newblock {\em bioRxiv\/}  (2021) 2021.04.23.441128.
\newblock \doi{10.1101/2021.04.23.441128}.
\newblock Publisher: Cold Spring Harbor Laboratory Section: New Results.

\bibitem[{Bellec et~al.(2018)Bellec, Salaj, Subramoney, Legenstein, and
  Maass}]{bellec_long_2018}
Bellec G, Salaj D, Subramoney A, Legenstein R, Maass W.
\newblock Long short-term memory and learning-to-learn in networks of spiking
  neurons.
\newblock {\em arXiv:1803.09574 [cs, q-bio]\/}  (2018).
\newblock ArXiv: 1803.09574.

\bibitem[{Frankle and Carbin(2018)}]{frankle_lottery_2018}
Frankle J, Carbin M.
\newblock The {Lottery} {Ticket} {Hypothesis}: {Training} {Pruned} {Neural}
  {Networks}  (2018).

\bibitem[{Willett et~al.(2021)Willett, Avansino, Hochberg, Henderson, and
  Shenoy}]{willett_high-performance_2021}
Willett FR, Avansino DT, Hochberg LR, Henderson JM, Shenoy KV.
\newblock High-performance brain-to-text communication via handwriting.
\newblock {\em Nature\/} {\bf 593} (2021) 249--254.
\newblock \doi{10.1038/s41586-021-03506-2}.
\newblock Number: 7858 Publisher: Nature Publishing Group.

\bibitem[{Jordan and Park(2020)}]{e22050537}
Jordan ID, Park IM.
\newblock Birhythmic analog circuit maze: A nonlinear neurostimulation testbed.
\newblock {\em Entropy\/} {\bf 22} (2020).
\newblock \doi{10.3390/e22050537}.

\bibitem[{Carr(1981)}]{carr_applications_1981}
Carr J.
\newblock {\em Applications of {Centre} {Manifold} {Theory}\/} (New York,
  Heidelberg, Berlin: Springer), 1982nd edition edn. (1981).

\end{thebibliography}

%%% Make sure to upload the bib file along with the tex file and PDF
%%% Please see the test.bib file for some examples of references

\end{document}